\documentclass[11pt,en]{elegantpaper}
\title{UdeerLID+: Integrating LiDAR, Image, and Relative Depth with Semi-Supervised}

\author{Tao Ni \\ UDeer.ai \\ nitao@udeer.ai \and Xin Zhan \\ UDeer.ai \\ zhanxin@udeer.ai \and Tao Luo \\ UDeer.ai \\ luotao@udeer.ai  \and Wenbin Liu \\ isep \\ wenbin@udeer.ai \and Zhan Shi \\ Zhejiang University \\ shizhan@udeer.ai \and JunBo Chen \\ UDeer.ai \\ junbo@udeer.ai}

\version{Technical Report}
\usepackage{array}

\begin{document}
\maketitle

\begin{abstract}
  Road segmentation is a critical task for autonomous driving systems, requiring accurate and robust methods to classify road surfaces from various environmental data. Our work introduces an innovative approach that integrates LiDAR point cloud data, visual image, and relative depth maps derived from images. 
  The integration of multiple data sources in road segmentation presents both opportunities and challenges. One of the primary challenges is the scarcity of large-scale, accurately labeled datasets that are necessary for training robust deep learning models. To address this, we have developed the [UdeerLID+] framework under a semi-supervised learning paradigm. 
  Experiments results on KITTI datasets validate the superior performance.
\end{abstract}

\section{Introduction}

The accurate identification of road areas in urban environments is a crucial capability for autonomous driving systems. Reliable road detection is essential for enabling autonomous vehicles to navigate safely and efficiently in complex real-world scenarios. While traditional vision-based road detection methods have seen substantial progress, they often face significant challenges such as varying illumination, shadow occlusion, and motion blur, which can compromise segmentation accuracy and reliability.

To overcome these challenges, we propose a novel method that integrates LiDAR data, relative depth, and visual imagery. LiDAR is particularly advantageous due to its resilience against visual noise and its ability to provide precise spatial information. Meanwhile, relative depth complements sparse LiDAR point clouds by offering additional spatial context. The fusion of these modalities—LiDAR, depth, and visual data—offers a comprehensive framework to improve road detection by leveraging their individual strengths.

Our method, Integrating LiDAR, Image, and Relative Depth with Semi-Supervised (UdeerLID+), introduces a two-step adaptation process to seamlessly integrate LiDAR information and Relative Depth into the visual domain. 
Like PLARD[1], the first step involves Data Space Adaptation, where raw LiDAR data is transformed into a format that aligns with the 2D perspective of visual data. This is achieved by utilizing an altitude difference-based transformation that preserves the distinguishing characteristics of road surfaces within the LiDAR data.
The robust monocular depth is generated by Depth Anything[2].

In the semi-supervised learning framework of [UdeerLID+], the model is trained using a combination of labeled and unlabeled data. The labeled data provides direct supervision, helping the model learn the mapping from input features to road segmentation labels. The unlabeled data, on the other hand, is used to learn generalizable representations that can improve the model's ability to generalize to new, unseen data.
In [3], the incorporation of meta-learning into the detection domain has inspired innovative frameworks for road segmentation tasks.

Extensive experiments on the KITTI road detection benchmark demonstrate that [UdeerLID+] surpasses state-of-the-art methods, significantly improving road detection accuracy across diverse urban conditions, including those characterized by challenging illumination.

In conclusion, [UdeerLID+] represents a substantial advancement in autonomous driving, offering a robust and precise solution for road detection by effectively integrating the complementary strengths of LiDAR, depth, and visual data.

\section{Related Work}

\subsection{Road Segmentation}
Road segmentation is a pivotal component for autonomous driving systems, focusing on the accurate identification and delineation of drivable road surfaces from environmental sensor data. This task is crucial for enabling vehicles to perceive their surroundings, make informed navigation decisions, and ensure safe path planning.
As PLARD[1], road segmentation is addressed by integrating and adapting LiDAR data with visual imagery to enhance the detection performance. The core objective of the road segmentation task is to assign binary labels to each pixel on the 2D image plane, distinguishing whether it corresponds to a road surface or a non-road area.

\subsection{Strong baseline: PLARD}
PLARD (Progressive LiDAR Adaptation-aided Road Detection[1] is designed for robust road segmentation in autonomous driving applications. It innovatively combines LiDAR data with visual imagery to enhance road detection accuracy and reliability. 
The method employs a two-step adaptation process: first, it adapts the LiDAR data into the visual data space using an altitude difference-based transformation to align the perspectives and preserve road features; second, it adapts the LiDAR features to the visual features through a cascaded fusion structure within a deep learning framework, allowing for a more comprehensive and accurate representation of the road environment. This dual-modal integration and progressive adaptation approach enable PLARD to achieve superior performance in road detection tasks, even under challenging conditions such as varying lighting or occlusions.

\subsection{Depth Anything}
Depth information, derived from various sensors such as LiDAR or stereo cameras, plays a crucial role in numerous applications across fields like autonomous driving, robotics, and augmented reality. 
DepthAnything[2] provides a quantitative measurement of the distance between the sensor and objects in the environment, enabling systems to perceive the three-dimensional structure of their surroundings. 
This understanding is vital for tasks such as obstacle detection, navigation, and scene reconstruction, where the precise estimation of depth helps in making informed decisions and actions based on the spatial relationships within the scene. 
Depth data also enhances the capabilities of machine learning models by offering an additional layer of insight that can improve the accuracy and reliability of predictive and analytical tasks.

\subsection{Meta Learning}
Meta learning[3], also known as "learning to learn" is a subfield of machine learning that focuses on developing algorithms that can efficiently learn from small amounts of data by leveraging prior knowledge or experience gained from training on related tasks. 
This approach enables models to quickly adapt to new problems with minimal data through techniques such as model agnostic meta learning (MAML), gradient-based meta learning, and reinforcement-based methods, thereby improving their generalization and flexibility across diverse tasks.

\subsection{Encoder Decoder}
Encoder-Decoder based segmentation models are a class of deep learning architectures designed for pixel-wise image segmentation tasks, where an encoder captures and compresses spatial information from the input image into a latent feature representation, and a decoder subsequently up-samples and maps this representation back to the original image resolution to produce a segmentation mask. 
These models[4][5][6] have been widely used in various segmentation domains, including semantic segmentation and instance segmentation, due to their ability to effectively model long-range dependencies in images. 

\subsection{Intern Image}
InternImage[7] is a novel large-scale convolutional neural network (CNN) foundation model that leverages deformable convolutions as its core operator, allowing it to dynamically adapt its receptive field and perform adaptive spatial aggregation. This model not only captures long-range dependencies essential for tasks like detection and segmentation but also reduces the strict inductive bias of traditional CNNs. By scaling up the model parameters and training data similar to Vision Transformers (ViTs), InternImage achieves state-of-the-art performance on various challenging benchmarks, including ImageNet, COCO, and ADE20K, demonstrating the potential of CNN-based models in the era of large-scale vision foundation models.

\section{METHOD}
\subsection{Overview}
Our UdeerLID+ is mainly construsted by two main components, multi-sources Encoder-Decoder based segmentator and Meta Pseudo Labels in the field of semantic segmentation. Both methods can effectively enhance performance metrics, and when employed in conjunction, they can lead to further improvements.
Below section would descript them in detail. The pipline of UdeerLID+ is shown in Fig. 1.
\begin{figure}[htbp]
\centering
\includegraphics[scale=0.3]{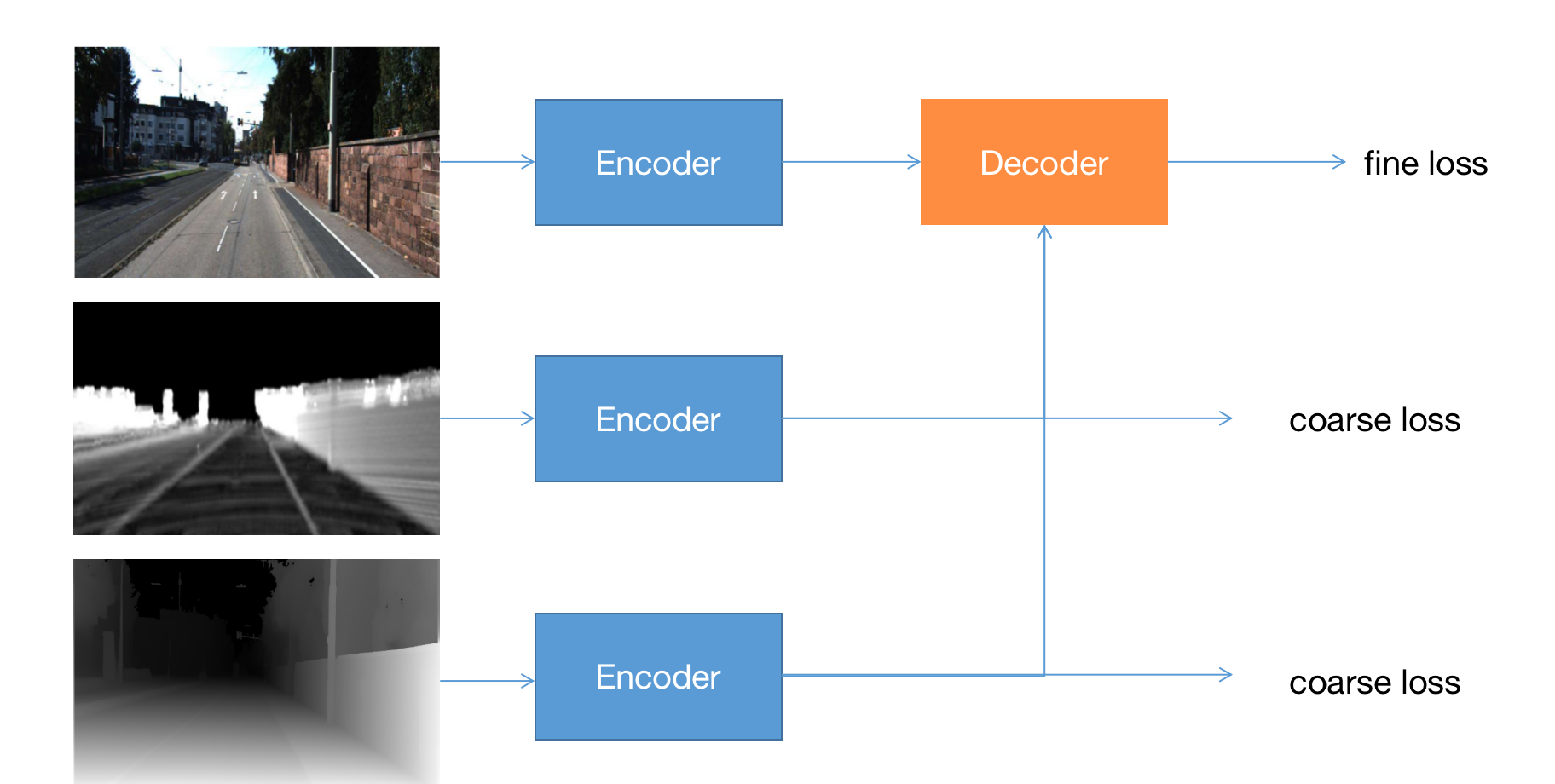}
\caption{UdeerLID+ Pipline}
\label{1}
\end{figure} 

\subsection{Multi-sources Encoder-Decoder}
Images offer rich texture and color information that is crucial for recognizing and distinguishing various objects and scenes, providing visual cues that are invaluable for tasks such as classification and object detection. 
LiDAR point clouds deliver precise three-dimensional spatial data with accurate distance measurements, making them robust against lighting variations and capable of capturing the geometric structure of the environment, which is essential for tasks like 3D object detection and segmentation. 
Relative depth maps, derived from images, provide an additional layer of depth perception that complements monocular vision by estimating the distance of objects within the scene, thus enhancing the understanding of the scene's layout and facilitating tasks such as depth estimation and semantic segmentation.

In our framework, image data, LiDAR point clouds, and relative depth maps synergistically contribute their unique strengths. 
Like Fig. 1, each modality has its dedicated encoder, which processes the input data and extracts distinctive features. Furthermore, an auxiliary loss function is applied at the output layer of each encoder to ensure that each modality can independently perform the road segmentation task ($Loss_{image}$, $Loss_{LiDAR}$, $Loss_{depth}$), thereby enhancing the overall performance and robustness of our multi-modal framework.
In the decoder phase of our framework, the image features take the central role, which are then augmented by integrating the upsampled features from LiDAR and depth data. This fusion process amalgamates the complementary strengths of all three sensor modalities, culminating in an enriched feature representation that encapsulates the multi-sensory information. 
The final segmentation $Loss_{fine}$  is computed on the merged features, and the total loss is formulated as follows: 

$$
Loss = Loss_{fine} + \alpha \cdot Loss_{image} + \beta \cdot Loss_{LiDAR} + \gamma \cdot Loss_{depth}
$$

Dettailed experimental result will be shown in next section for multi-modality road estimation.

\subsection{Meta Pseudo Labels}
The KITTI road segmentation dataset, while limited to just over 200 training images, is complemented by a substantial collection of unlabeled yet thematically consistent data within the KITTI detection dataset. 
This extensive repository of similar unlabeled images serves as a rich source for semi-supervised learning, allowing models to generalize better and improve their segmentation capabilities without the need for extensive manual annotation.

In the KITTI road segmentation task, we employ a semi-supervised learning approach utilizing meta learning techniques. The process begins with the training of a fully supervised model on the complete dataset, which is represented as $Model_{Supervised}$.

Following the initial training, we commence the semi-supervised learning phase with an iterative algorithm. For each iteration, we define a confidence threshold \( \tau \) for the model's predictions, and only pixels with confidence above this threshold contribute to the loss function, while others are masked. This can be mathematically expressed as:

\[
\text{Semi-Supervised Update} = arg_{min}{\theta} \sum_{i \in S} \mathcal{L}(y_i, \hat{y}_i; \theta),
\]

where \( \theta \) denotes the model parameters, \( \mathcal{L} \) is the loss function, \( y_i \) are the pseudo labels, and \( \hat{y}_i \) are the predicted labels.
\( S \) is the subset of pixels with confidence scores above the threshold \( \tau \), and the remaining pixels are masked to prevent them from influencing the learning process.

This method allows the model to progressively refine its understanding of the road segmentation task by focusing on high-confidence predictions and incorporating unlabeled data in a controlled manner.

\section{Experiments}
\subsection{Dataset}
We validate the proposed UdeerLID+ on KITTI dataset.

The KITTI road dataset, is a comprehensive benchmark designed for evaluating road segmentation algorithms in the context of autonomous driving. 
It features a diverse set of real-world driving scenarios, captured by high-resolution cameras and LiDAR sensors mounted on a moving vehicle. 
The dataset consists of over 200 training images and 200 test images, each paired with accurate ground truth annotations that delineate the drivable road regions. 
The KITTI road dataset serves not only as a testbed for algorithm development but also as a valuable resource for advancing the state-of-the-art in road detection technology.

\subsection{Implementation Details}
In the implementation, an encoder-decoder architecture is utilized, specifically using the InternImage-B model[7] pretrained on the ADE20K Semantic Segmentation dataset. 
This pretrained model serves as the backbone for the image encoder, effectively capturing and encoding the semantic information from the input images. 
The decoder then upscales these encoded features to generate high-resolution segmentation maps, providing detailed and accurate delineation of road regions.

\subsection{Connection}
In the integration of images, point clouds, and depth information, our approach draws inspiration from PLARD[1], employing a multi-level fusion and upsampling strategy. 
This method involves the synergistic combination of these diverse data types at various stages of the network. By conducting feature fusion at multiple hierarchical levels, we enhance the representational power of the model, which is crucial for tasks such as semantic segmentation and object detection that require detailed understanding of the scene's geometry and appearance.

\subsection{Quantitive Evaluation}
The quantitive results on KITTI dataset are shown in Table 1. PLARD's result is based on its released model on the Github; UdeerLID is the version without Meta Pseudo Labels.
\begin{table}[h!]
  \begin{center}
    \caption{Um Road Estimation On The Test Set Of KITTI Dataset}
    \begin{tabular}{l|c|r} % <-- Alignments: 1st column left, 2nd middle and 3rd right, with vertical lines in between
      & method & MaxF \\
      \hline
      1 & PLARD[1] & 96.32\\
      2 & UdeerLID & 96.94\\
      3 & UdeerLID+ & 97.26\\
    \end{tabular}
  \end{center}
\end{table}
\subsection{Discussion}
The effectiveness of relative depth in the task of road recognition has been demonstrated, showcasing its potential to enhance the accuracy and robustness of road detection models. 
Additionally, the integration of meta-learning techniques in semantic segmentation has been observed to significantly contribute to the growth of model performance. By leveraging the unique advantages of each data modality and employing advanced learning strategies, we can achieve more reliable and efficient models for autonomous driving and computer vision applications.

\section{Future Work}
This endeavor aims to optimize model performance by exploring various combinations of visual imagery, LiDAR point clouds, relative depth maps, and other relevant sensor data. The goal is to identify the optimal data fusion architectures and algorithms that can best capture the complexities of the driving environment, ultimately leading to more reliable and intelligent autonomous systems.

\appendix
%\appendixpage
\addappheadtotoc

\end{document}